\def\year{2018}\relax
\documentclass[letterpaper]{article} 
\usepackage{aaai18}  
\usepackage{times}  
\usepackage{helvet}  
\usepackage{courier}  
\usepackage{url}  
\usepackage{graphicx}  

\graphicspath{ {figs/} }
\usepackage[FIGTOPCAP]{subfigure} 

\usepackage{algorithm}
\usepackage{algorithmic}
\usepackage{amssymb}
\usepackage{amsmath}
\usepackage{textcomp}
\usepackage{multirow}
\usepackage{color}

\usepackage[inline]{enumitem}

\usepackage[utf8]{inputenc} 
\usepackage[T1]{fontenc}    
\usepackage{url}            
\usepackage{booktabs}       
\usepackage{amsfonts}       
\usepackage{nicefrac}       
\usepackage{microtype}      

\title{Lattice Recurrent Unit:\\ Improving Convergence and Statistical Efficiency for Sequence Modeling}
\author{Chaitanya Ahuja\\
  Language Technologies Institute\\
  Carnegie Mellon University\\
  Pittsburgh, PA 15213 \\
  \texttt{cahuja@andrew.cmu.edu} \\
  \And
  Louis-Philippe Morency \\
  Language Technologies Institute\\
  Carnegie Mellon University\\
  Pittsburgh, PA 15213 \\
  \texttt{morency@cs.cmu.edu} \\
}

\begin{document}
%

\maketitle
\begin{abstract}
  Recurrent neural networks have shown remarkable success in modeling sequences. However low resource situations still adversely affect the generalizability of these models. We introduce a new family of models, called Lattice Recurrent Units (LRU), to address the challenge of learning deep multi-layer recurrent models with limited resources.
  LRU models achieve this goal by creating distinct (but coupled) flow of information inside the units: a first flow along time dimension and a second flow along depth dimension. It also offers a symmetry in how information can flow horizontally and vertically.
  We analyze the effects of decoupling three different components of our LRU model: \textit{Reset Gate}, \textit{Update Gate} and \textit{Projected State}. We evaluate this family of new LRU models on computational convergence rates and statistical efficiency.
Our experiments are performed on four publicly-available datasets, comparing with Grid-LSTM and Recurrent Highway networks. Our results show that LRU has better empirical computational convergence rates and statistical efficiency values, along with learning more accurate language models. 
\end{abstract}

\section{Introduction}
Recurrent Neural Networks have been shown to be turing complete \cite{siegelmann1995computational} and hence can approximate any given function. Even though they can theoretically represent any form of sequential data, these networks are hard to optimize by gradient methods as gradients start diminishing if backpropagated over large number of time-steps \cite{bengio1994learning,pascanu2013difficulty,hochreiter2001gradient,pascanu2013difficulty}. This is overcome by the use of gating mechanisms in Long Short-Term Memory (LSTMs) \cite{hochreiter1997long} and more recently Gated Recurrent Units (GRUs) \cite{cho2014properties}. 
\begin{figure}[ht]
\centering
\includegraphics[width=1\linewidth]{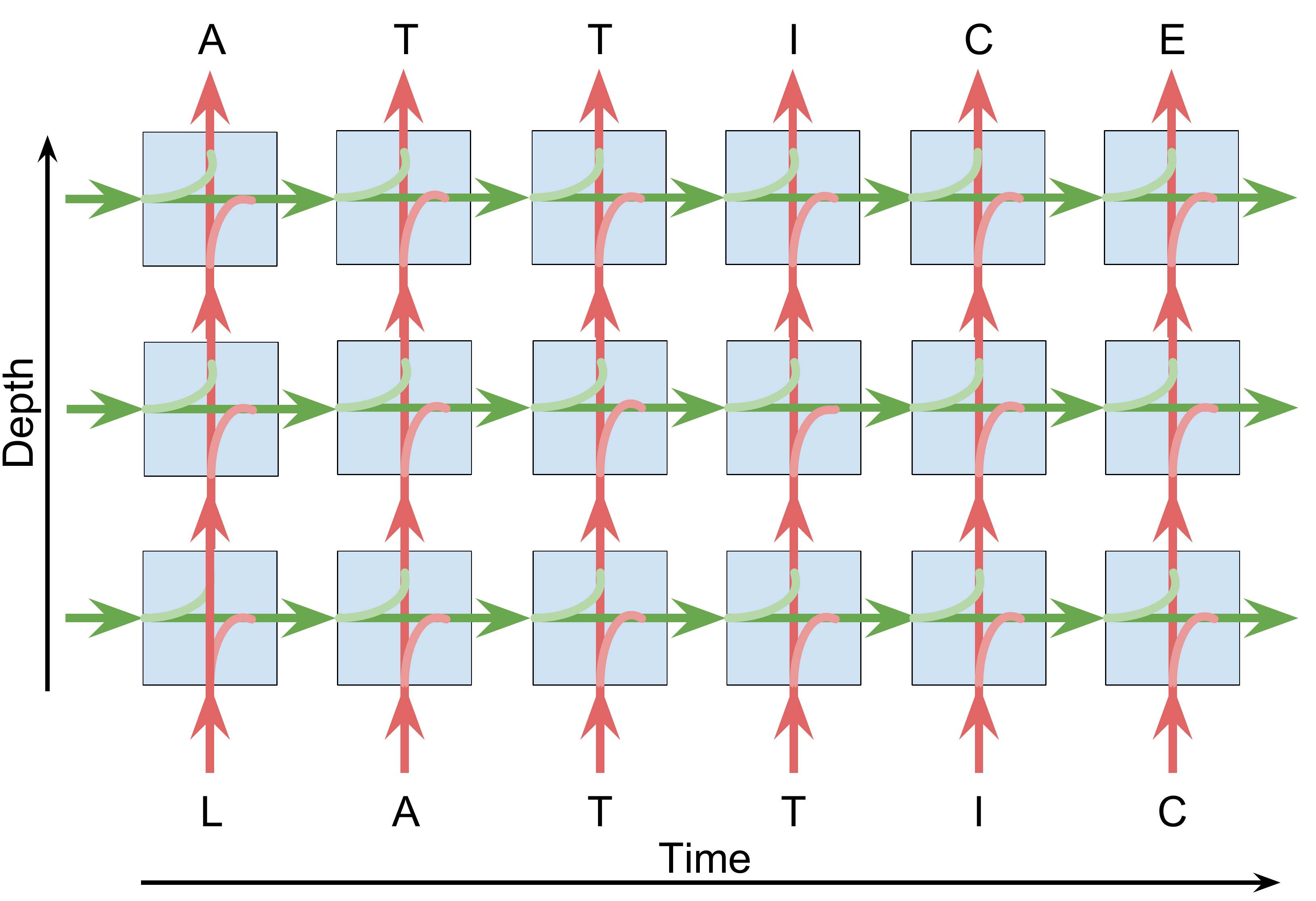}
\caption{A depiction of how our proposed Lattice Recurrent Unit (LRU) can create a distinct flows of information for both time and depth dimensions. LRU enables better convergence and accuracy, especially with limited resource settings. We exemplify LRU with a character prediction task. The figure shows LRU predicting the word "LATTICE", using the "L" character as the starting point.}
\label{fig:overview}
\end{figure}
Gates ensures a constant flow of backpropagated error along the temporal dimension, hence making neural sequence models trainable (or optimizable). GRUs, LSTMs and variants of gated recurrent networks seem to capture temporal dependencies and have been successful in tasks such as language modeling \cite{jozefowicz2016exploring}, machine translation \cite{sutskever2014sequence}, handwriting recognition \cite{graves2009offline} and generation \cite{graves2013generating}, image captioning \cite{vinyals2014show} and video captioning \cite{venugopalan2015sequence}.

The depth of a neural network makes modeling of the data exponentially more efficient \cite{bianchini2014complexity}, but it is a challenge to optimize the parameters of multi-layer (or deep) models, especially under low resource settings. GridLSTM \cite{kalchbrenner2015grid} and Recurrent Highway Networks (RHN) \cite{zilly2016recurrent} were introduced to improve training of deep LSTM models as it is impractical to train very deep stacked LSTM or GRU models due to the vanishing and exploding gradient problem along depth.

\begin{figure*}[ht]
\centering
\subfigure[GRU]{\includegraphics[width=0.25\textwidth]{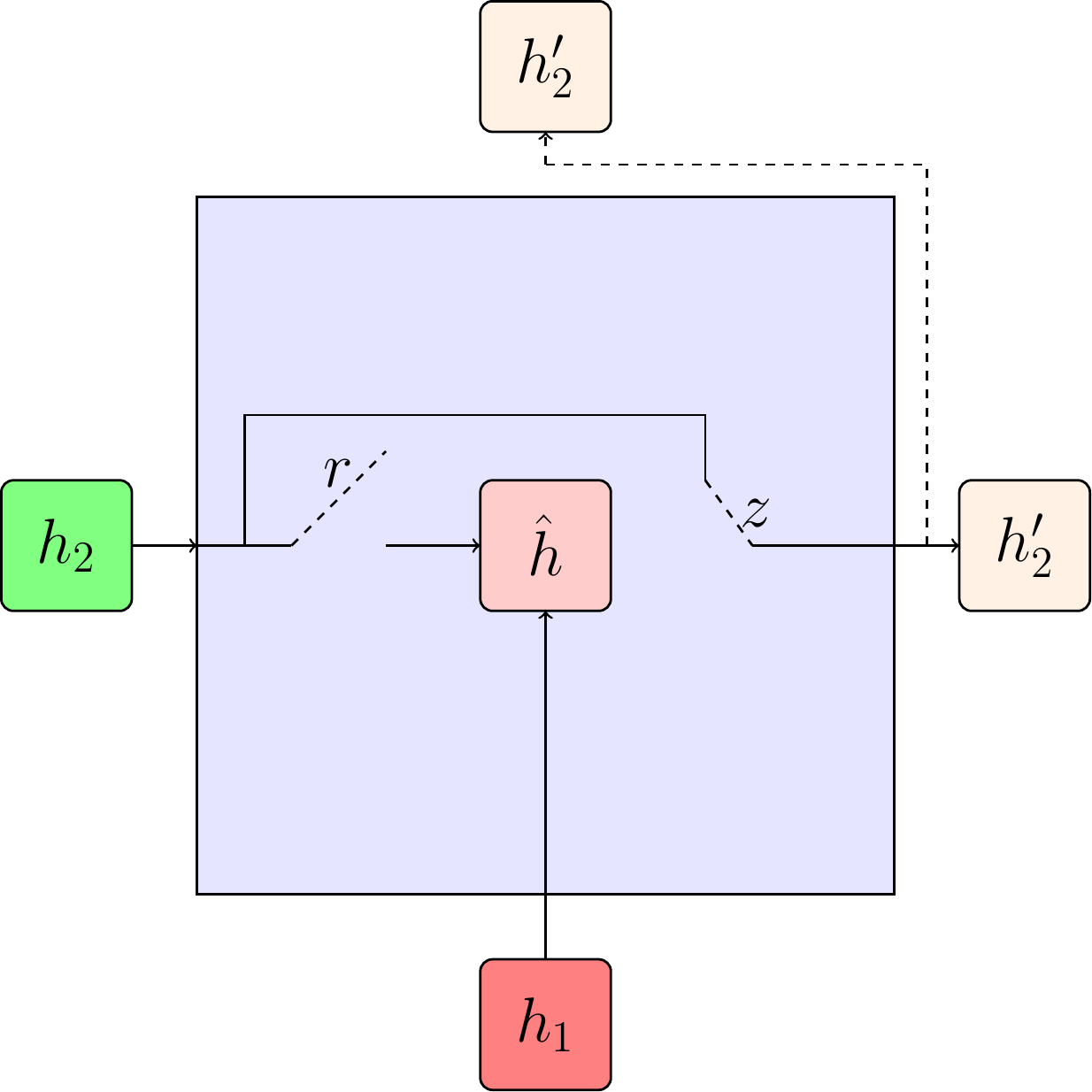}}
\hspace{1cm}
\subfigure[LRU]{\includegraphics[width=0.25\textwidth]{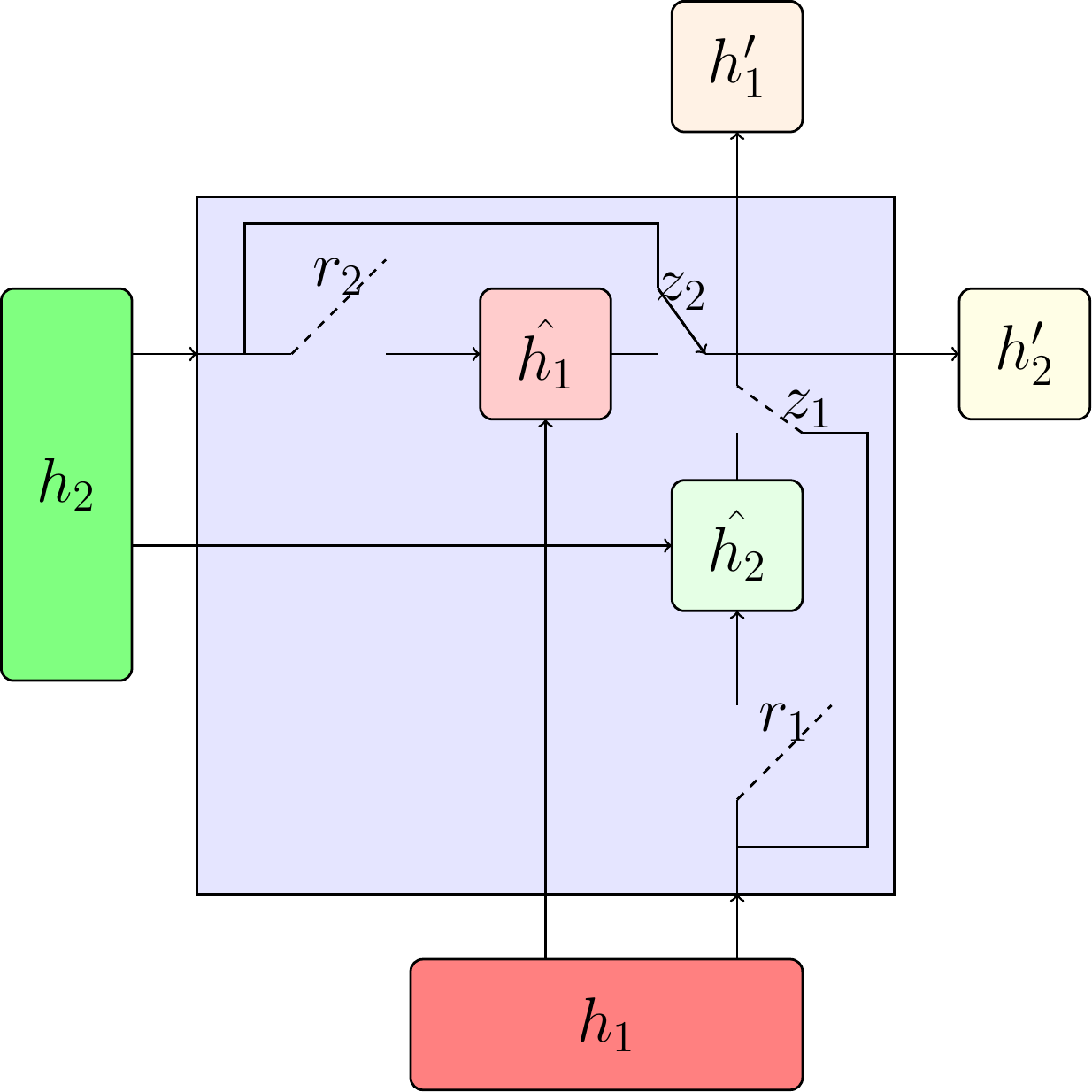}}
\hspace{1cm}
\subfigure[GridLSTM]{\includegraphics[width=0.25\textwidth]{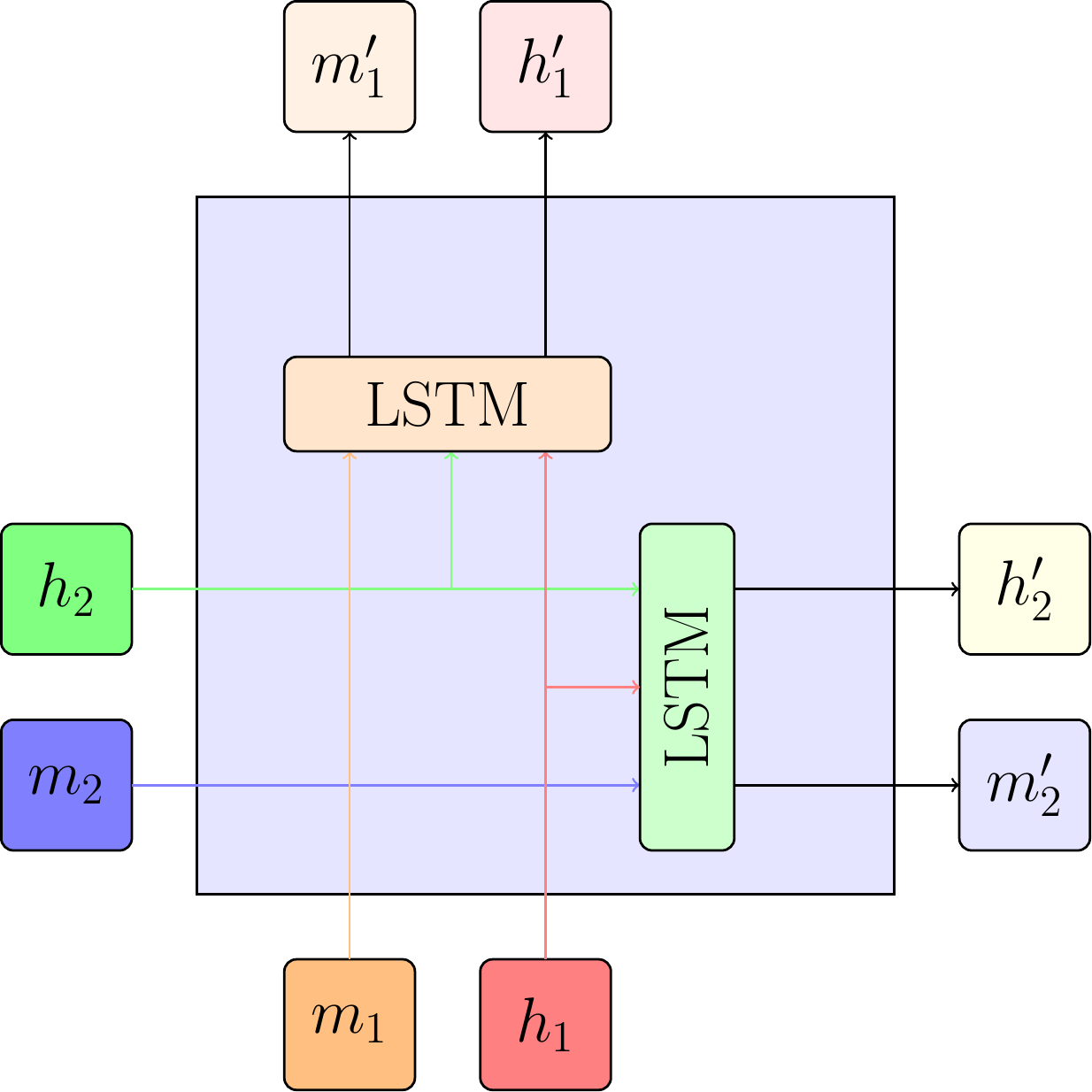}}

\caption{Comparison of our Lattice Recurrent Unit (LRU) model (shown in the middle) compared with the simpler Gated Recurrent Unit (GRU) model and the more recent GridLSTM model. LRU units create a distinct flow of information along  time and depth dimensions unlike a GRU unit. We also propose two more variants of our model: Reset Gate LRU (RG-LRU) which couples the update gates ($z_1$ and $z_2$), and the Projected State LRU (PS-LRU) which couples the Reset Gates ($r_1$ and $r_2$). Decoupling gates in the LRU model allows adaptation to different information propagated along depth and time which gives a boost in accuracy, convergence rates and statistical efficiency.}

\label{fig:compare}
\end{figure*}

In this paper we propose a family of models, called Lattice Recurrent Units (LRUs), to address the challenge of deep recurrent models. LRU and its variants (Projected State LRU, Reset Gate LRU) are an adaptation of GRUs to a lattice multi-dimensional architecture. The structural differences amongst the variants lies in coupling (or decoupling) of some or all weights. To observe the effects of decoupling weights, we perform experiments on four language modeling datasets and compare performancess of GRUs to all the proposed variants of LRUs. As there has been an increasing interest in training speeds of RNNs by parallelizing computations \cite{bradbury2016quasi,vaswani2017attention} and reducing complexity models \cite{joulin2016bag}, we also compare all the models with \textit{computational convergence rates} and \textit{statistical efficiency} \cite{bojanowski2016enriching} as metrics. Also, as a comparison to state-of-the art recurrent units, we perform the same set of experiments on LSTMs, GridLSTMs and RHNs.

\section{Background}
This section introduces the technical background related to our new Lattice Recurrent Unit (LRU) models. First, we describe the Gated Recurrent Units (GRUs) \cite{cho2014properties} and their multi-layer extensions called Stacked GRU. Our LRU models builds upon this family of recurrent units to enable deeper representations. Second, we describe Long Short-Term Memory (LSTM) \cite{hochreiter1997long} models and their multi-layer extensions, specifically the Grid LSTM model \cite{kalchbrenner2015grid} and Recurrent Highway Networks \cite{zilly2016recurrent} given their relevance to our research goals. These models will also be used in our experiments as our baselines.

\subsection{Notation for Sequence Modeling}
Consider an ordered input-output sequence $\lbrace(x^{(i)},y^{(i)})\rbrace_N$ of length N where $x^{(i)} \in \mathbb{R}^d$ is the input, while $y^{(i)} \in \mathbb{R}^d$ is the corresponding output. By an ordered sequence, we mean that $x^{(i+1)}$ comes after $x^{(i)}$ in time. Language modeling is a classic example where the inputs could be characters (or words) and the sequence is a sentence. The order of the characters (or words) is important to retain the meaning of the sentence. Hence, to model the dependencies of a sample $(x^{(n)},y^{(n)})$, it would be useful to have some information from the previous time-step samples $\lbrace(x^{(n-1)},y^{(n-1)}), (x^{(n-2)},y^{(n-2)})\ldots (x^{(1)},y^{(1)})\rbrace$.
\subsection{Gated Recurrent Unit}
The original Recurrent Neural Networks (RNNs) \cite{elman1990finding,rumelhart1988learning} were designed to model sequential data but it can be hard to optimize its parameters \cite{pascanu2013difficulty,bengio1994learning}. Some gated variations were proposed\cite{yao2015depth,chung2015gated,greff2016lstm} to counter the effects of vanishing gradients. For the purpose of this paper, we will consider GRU as described in \cite{cho2014properties}.

The output of a GRU is also called \textit{hidden state} which is encoded with memory of the sequence till the current time step. GRUs transform the \textit{current input} $x^{(n)}$ and input \textit{hidden state} $h^{(n-1)}\in \mathbb{R}^m$ to $h^{(n)}$ (output \textit{hidden state}) which is used as implicit memory for the next input $x^{(n+1)}$. $m$ is the \textit{hidden state}'s size.

For consistency with notation from later part of this paper, we define \textit{hidden state} flow through horizontal direction (or time dimension) as $h_2$. Similarly, \textit{hidden state} flow  through vertical direction (or depth dimension) is defined as $h_1$. In case of a GRU model, $h_2$ represents input hidden state and $h'_2$ represents output \textit{hidden state}. $x^{(n)}$ is seen as a vertical input, which means that $h_1=x^{(n)}$. It's important to note that GRU does not have a dedicated output for the vertical dimension. Multi-layer GRUs simply replicate the horizontal output for the vertical dimension: $h'_2=h'_1$. Formulation of each unit is as follows (biases have been excluded for brevity).
 
\begin{eqnarray}
z &=&  \sigma\left( \begin{bmatrix}
  \mathbf{W^z} & \mathbf{U^z} 
\end{bmatrix}
\begin{bmatrix}
  h_1 \\
  h_{2}
\end{bmatrix} \right) \label{eq:1}\\
r &=&  \sigma\left( \begin{bmatrix}
  \mathbf{W^r} & \mathbf{U^r} 
\end{bmatrix}
\begin{bmatrix}
  h_1 \\
  h_{2}
\end{bmatrix} \right)\label{eq:2}\\
\hat{h} &=&  \tanh\left( \begin{bmatrix}
  \mathbf{W^h} & \mathbf{U^h} 
\end{bmatrix}
\begin{bmatrix}
  h_1 \\
  r \odot h_{2}
\end{bmatrix} \right)\label{eq:3}\\
h_2' &=& z \odot h_{2} + (1-z) \odot \hat{h} \label{eq:4}
\end{eqnarray}

where, $\mathbf{W^z},\mathbf{W^r},\mathbf{W^h}$ and $\mathbf{U^z},\mathbf{U^r}, \mathbf{U^h}$ are transform matrices for input and hidden states. $\sigma$ and $\tanh$ are the usual logistic sigmoid and hyperbolic tangent functions respectively, while $\odot$ is element-wise product. 

Gating unit $r$ is often called \textit{Reset Gate} and controls the flow of information coming from input \textit{hidden state} $h_{2}$. This transformation decides which features from the previous hidden state, alongside input state $h_1$, are projected onto a common space to give $\hat{h}$. We refer to this common space $\hat{h}$ as (\textit{Projected State}). Gating unit $z$ is often referred to as \textit{Update Gate} and decides the fraction of $h_2$ and $\hat{h}$ that is passed to the next time step.

To increase the capacity of GRU networks ~\cite{hermans2013training}, recurrent layers can be stacked on top of each other.
Since GRU does not have two output states, the same output \textit{hidden state} $h_2'$ is passed to the next vertical layer. In other words, the $h_1$ of the next layer will be equal to $h_2'$. This forces GRU to learn transformations that are useful along depth as well as time. While the model can be potentially trained to learn such transformations with a very large training set. When the training set size is limited, a more natural way would be to have two output states, one for the time dimension and a second for the depth dimension.


\subsection{Long Short-Term Memory}
Long Short-Term Memory (LSTM) models were introduced in 1997 to address the issue of vanishing gradient with recurrent neural networks~\cite{hochreiter1997long}. GRUs and LSTMs have been compared extensively \cite{zaremba2015empirical,greff2016lstm} on numerous tasks. GRUs performances are generally on par with LSTMs. The major difference in formulations of GRUs and LSTMs is the presence of dedicated \textit{Memory State} in an LSTM cell, unlike GRUs which have \textit{Memory State} encoded in the \textit{Hidden State} itself.




\subsection{Grid-LSTM}
Grid-LSTM \cite{kalchbrenner2015grid} can produce N distinct \textit{Memory States} as outputs where $N \geq 1$. To help understand the Gird-LSTM model, let's consider a unit in two dimensional configuration (i.e. N=2; one dimension for time and other for depth). Borrowing notation from \cite{kalchbrenner2015grid}, we can write:
\begin{eqnarray}
h'_{1},m'_{1} = \mbox{\textbf{LSTM}}(h_{1},h_{2},m_{1},W_{1}) \label{eq:grid1}\\
h'_{2},m'_{2} = \mbox{\textbf{LSTM}}(h_{1},h_{2},m_{2},W_{2}) \label{eq:grid2}
\end{eqnarray}
where ($h_1, h_2$) and ($m_1,m_2$) are input \textit{hidden} and \textit{memory} states while($h_1', h_2'$) and ($m_1',m_2'$) are output \textit{hidden} and \textit{memory} states. $W_1$ and $W_2$ are the set of parameters for each dimension, where the subscripts denote the direction of flow of information.  $\textbf{LSTM}$ function is the typical LSTM recurrent unit described in~\cite{hochreiter1997long}. A Grid-LSTM unit spits out 2 distinct hidden and \textit{Memory States}, each of which is fed to different dimensions. Both time and depth have their very own forget gates, and this clever trick allows for reduction in the effect of vanishing gradient during optimization.


\subsection{Recurrent Highway Network}
Recurrent Highway Networks (RHN) \cite{zilly2016recurrent} increase the number of non-linear transformations in one recurrent unit to increase the capacity of the model. This combined with a gated sum of previous and current hidden states \cite{srivastava2015highway} ensures trainability of the resulting network, even if it were very deep. Even though, RHNs have more capacity they lack the ability of transferring intermediate hidden states along depth to the subsequent time step.


\section{Lattice Recurrent Unit}
In this section we introduce a family of models, Lattice Recurrent Units, designed to have distinct flow of information for through time and depth dimensions. LRU models can be seen as an expansion of the GRU model. As shown in Equations (\ref{eq:1}-\ref{eq:3}), GRU has three main components: \textit{Projected State} $\hat{h}$, \textit{Reset Gate} $r$ and \textit{Update Gate} $z$. In this paper, we created three members of the LRU family to study an important research question: what components of the GRU model should be decoupled to enable two channels of information?

The first LRU model, named \textit{Projected State} LRU (PS-LRU) will decouple only \textit{projected states} for each dimension. The second model, named \textit{Reset Gate} LRU (RG-LRU), will go one step further by also decoupling \textit{reset gates}. Finally the \textit{Update Gate} LRU (UG-LRU), which we also call LRU, will decouple all three components, including the \textit{update gate}.
We decouple one gate at a time and formulate 3 different members of LRU family. The following subsections describe each LRU model in more detail.


\subsection{Projected State LRU (PS-LRU)}
As a first model PS-LRU, we decouple \textit{projected state} $\hat{h}$ to create two new projected states: $\hat{h}_1$ and $\hat{h}_2$. Each of them is used to compute a separate output state: $h_1'$ and $h_2'$. Formally, PS-LRU is defined using the following update functions,


\begin{eqnarray}
\hat{h}_1 &=&  \tanh\left( \begin{bmatrix}
  \mathbf{W^h_1} & \mathbf{U^h_1} 
\end{bmatrix}
\begin{bmatrix}
  h_1 \\
  r \odot h_{2}
\end{bmatrix} \right)\label{eq:lru1}\\
\hat{h}_2 &=&  \tanh\left( \begin{bmatrix}
  \mathbf{W^h_2} & \mathbf{U^h_2} 
\end{bmatrix}
\begin{bmatrix}
  r \odot h_1 \\
  h_{2}
\end{bmatrix} \right)\label{eq:lru2}\\
  h_1' &=& z \odot h_{1} + (1-z) \odot \hat{h}_2  \label{eq:lru3}\\
  h_2' &=& z \odot h_2 + (1-z) \odot \hat{h}_1 \label{eq:lru4}
\end{eqnarray}

PS-LRU model uses the same \textit{update} and \textit{reset} gates for both output states (see Equation \ref{eq:1} and \ref{eq:2}). Note that Equation \ref{eq:4} splits up into Equations \ref{eq:lru3} and \ref{eq:lru4} as there are two distinct outputs in this model.

\subsection{Reset Gate LRU (RG-LRU)}

We go one step further and decouple the \textit{Reset Gate} (which was originally defined in Equation \ref{eq:2}) to give us two new gates, $r_1$ and $r_2$. This model is called RG-LRU and is defined as follows,
\begin{eqnarray}
r_1 &=&  \sigma\left( \begin{bmatrix}
  \mathbf{W^r_1} & \mathbf{U^r_1} 
\end{bmatrix}
\begin{bmatrix}
  h_1 \\
  h_{2}
\end{bmatrix} \right)\label{eq:lru5}\\
r_2 &=&  \sigma\left( \begin{bmatrix}
  \mathbf{W^r_2} & \mathbf{U^r_2} 
\end{bmatrix}
\begin{bmatrix}
  h_1 \\
  h_{2}
\end{bmatrix} \right)\label{eq:lru6}\\
\hat{h}_1 &=&  \tanh\left( \begin{bmatrix}
  \mathbf{W^h_1} & \mathbf{U^h_1} 
\end{bmatrix}
\begin{bmatrix}
  h_1 \\
  r_2 \odot h_{2}
\end{bmatrix} \right)\label{eq:lru9}\\
\hat{h}_2 &=&  \tanh\left( \begin{bmatrix}
  \mathbf{W^h_2} & \mathbf{U^h_2} 
\end{bmatrix}
\begin{bmatrix}
  r_1 \odot h_1 \\
  h_{2}
\end{bmatrix} \right)\label{eq:lru10}
\end{eqnarray}

Note that Equation \ref{eq:3} splits up into Equations \ref{eq:lru9} and \ref{eq:lru10} as $r$ has now been decoupled.

\subsection{Update Gate LRU (UG-LRU or LRU)}
In our final model, UG-LRU (which we also refer as LRU for simplicity), we decouple all three main components of GRU, including \textit{Update Gate} (originally defined in Equation \ref{eq:1}) which give us gates $z_1$ and $z_2$ defined as:
\begin{eqnarray}
z_1 &=&  \sigma\left( \begin{bmatrix}
  \mathbf{W^z_1} & \mathbf{U^z_1} 
\end{bmatrix}
\begin{bmatrix}
  h_1 \\
  h_{2}
\end{bmatrix} \right)\label{eq:lru7}\\
z_2 &=&  \sigma\left( \begin{bmatrix}
  \mathbf{W^z_2} & \mathbf{U^z_2} 
\end{bmatrix}
\begin{bmatrix}
  h_1 \\
  h_{2}
\end{bmatrix} \right)\label{eq:lru8}\\
h_1' &=& z_1 \odot \hat{h}_2 + (1-z_1) \odot h_{1} \label{eq:lru11}\\
h_2' &=& z_2 \odot \hat{h}_1 + (1-z_2) \odot h_2 \label{eq:lru12}
\end{eqnarray}
Note that Equations \ref{eq:lru3} and \ref{eq:lru4} needed a slight modification as $z$ has now been decoupled.

\begin{figure*}[ht]
\subfigure[\texttt{PTB}]{\includegraphics[width=0.24\textwidth]{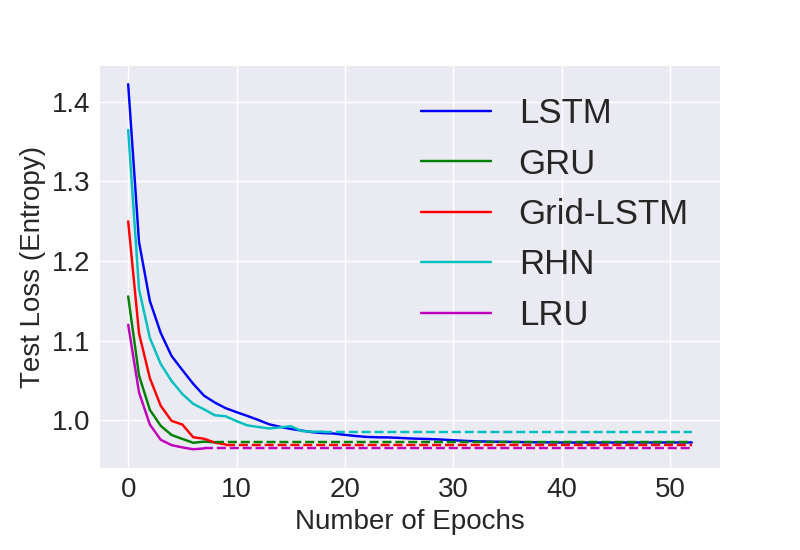}}
\subfigure[\texttt{WP}]{\includegraphics[width=0.24\textwidth]{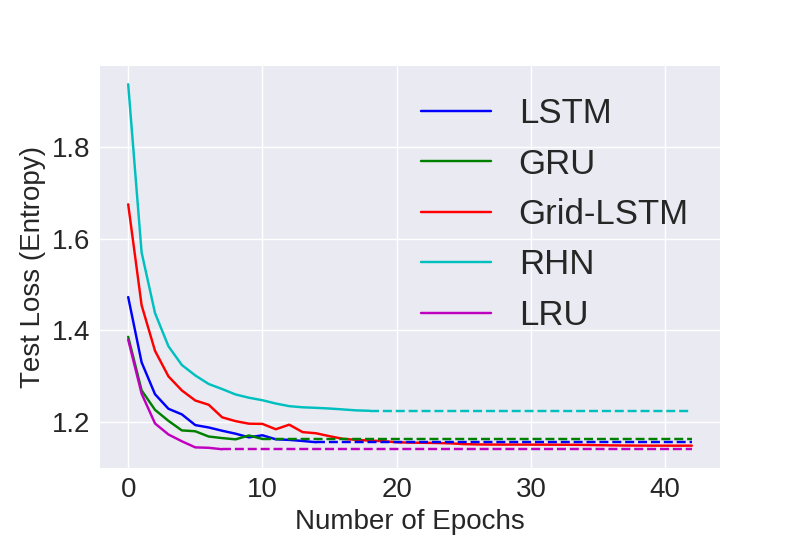}}
\subfigure[\texttt{enwik8}]{\includegraphics[width=0.24\textwidth]{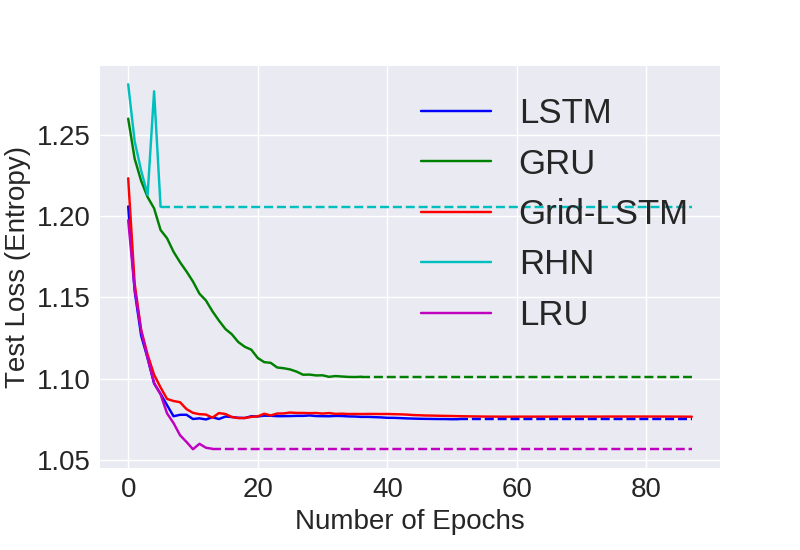}}
\subfigure[\texttt{text8}]{\includegraphics[width=0.24\textwidth]{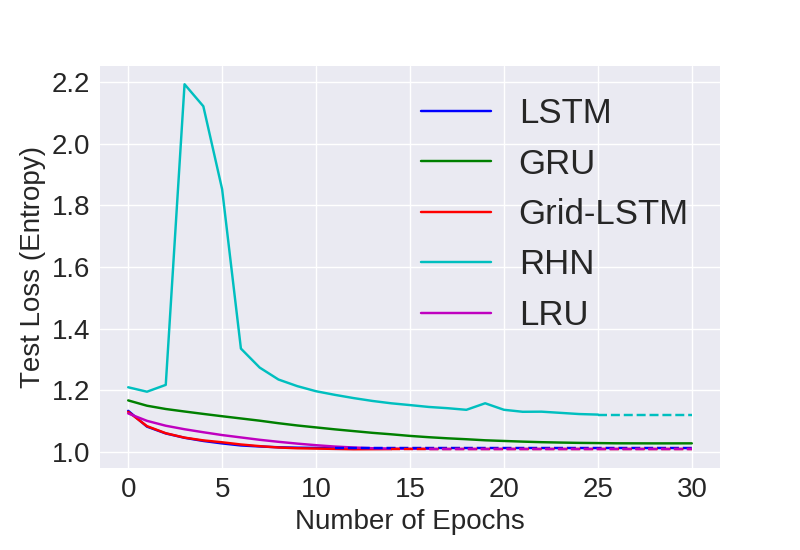}}

\caption{Visual representation of computational convergence rates of LSTM, GRU, RHN, Grid-LSTM and LRU on datasets: \texttt{PTB}, \texttt{WP}, \texttt{enwik8} and \texttt{text8}. Solid lines denote that actual training curves, while dotted lines are just an extension to visually compare the test score. For example in Figure (a) LRU training ends at around 10 epochs, while LSTM's training continues on till around 50 epochs. LRU converges \textit{faster} to a \textit{lower CCE} loss on unseen data than the other models. All the models have 10M parameters\protect\footnotemark. Y-axis is loss as Categorical Crossentropy Error (CCE) on test sets, while X-axis is number of epochs.}
\label{fig:test_curves}
\end{figure*}

\begin{figure*}[ht]
\subfigure[\texttt{PTB}]{\includegraphics[width=0.24\textwidth]{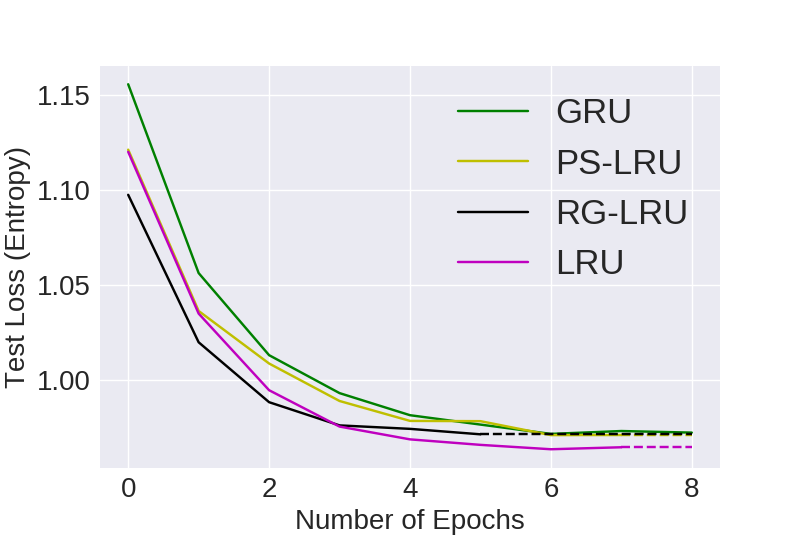}}
\subfigure[\texttt{WP}]{\includegraphics[width=0.24\textwidth]{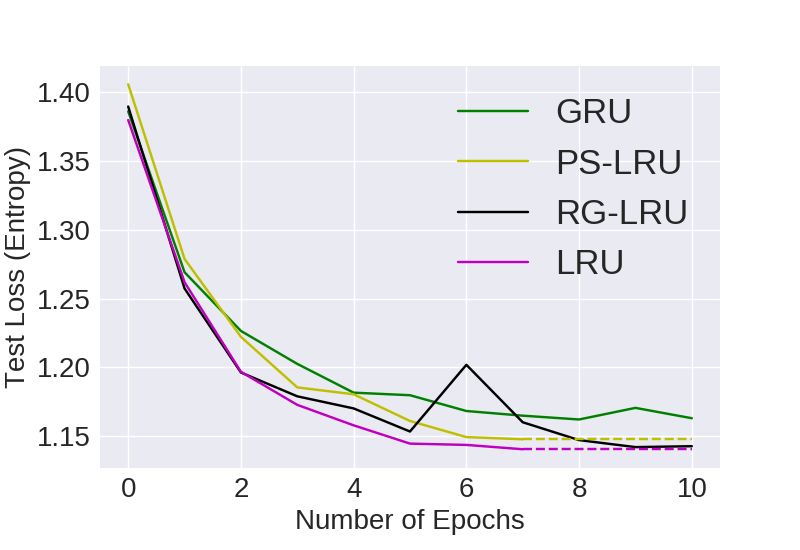}}
\subfigure[\texttt{enwik8}]{\includegraphics[width=0.24\textwidth]{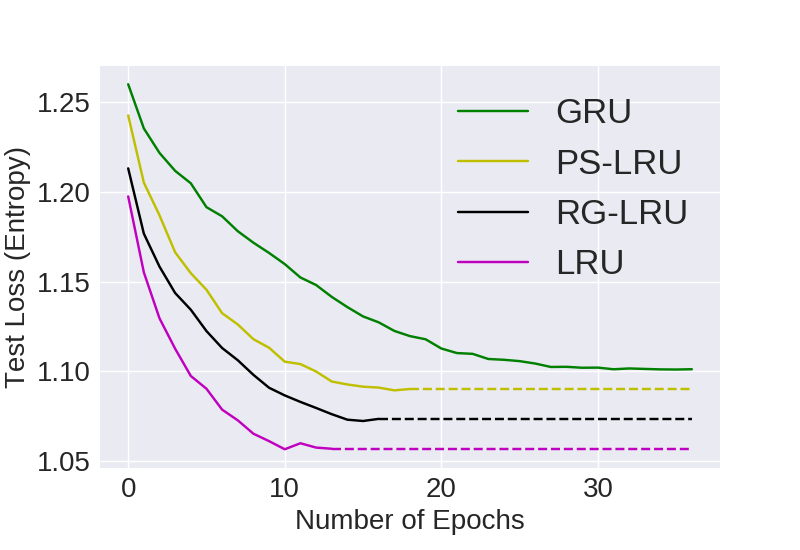}}
\subfigure[\texttt{text8}]{\includegraphics[width=0.24\textwidth]{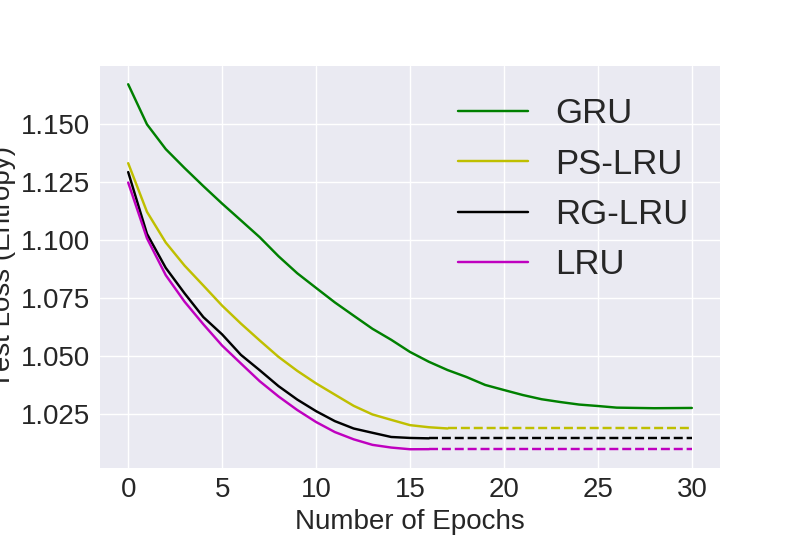}}

\caption{Visual representation of computational convergence rates of GRU, PS-LRU, RG-LRU and LRU on datasets: \texttt{PTB}, \texttt{WP}, \texttt{enwik8} and \texttt{text8}. LRU and GRU are at two extreme ends of the coupled-weights spectrum. These training curves show that systematically decoupling weights in a GRU (to formulate PS-LRU, RG-LRU and eventually LRU) gives significant boosts to the \textit{convergence rate} and \textit{accuracy} of the model. All the models have 10M parameters.}
\label{fig:test_curves_comp}
\end{figure*}

\section{Experiments}
One of the objectives is to study the effects of coupling gates and control flow of information, hence we split experiments into two parts. The first one is a comparative study on \textbf{GRU}  and \textbf{LRU} family (\textbf{PS-LRU}, \textbf{RG-LRU}, \textbf{LRU}), while the second one compares \textbf{LRU} with baseline models including \textbf{LSTM}, \textbf{GRU}, \textbf{RHN} and \textbf{Grid-LSTM}\footnote{In a Grid-LSTM model it is possible to couple weights across all or some dimensions, but for the sake of comparison we couple the weights across depth.}. Our LRU models and all baseline models were all trained and tested with the same methodology. In other words, we are not copying result numbers from other papers but instead testing every model to be sure that we have fair comparisons. This focus on fair comparisons (in contrast with other approach to only focus on state-of-the-art performance, even if the experimental environment is different) motivates our choice to not include dropout or variational layers in any of the models.


Models are compared on 3 evaluation metrics,
\begin{enumerate}[label={(\arabic*)}]
\item \textbf{Accuracy} - Categorical Cross Entropy (CCE) is used as the loss function and is also used to compare all the models. Lower is better. Model with the best validation loss is used to compute this loss.
\item \textbf{Computational Convergence Rate} - Number of epochs (traversal over complete dataset) taken to converge to the best possible model based on validation scores. Lesser number of epochs to complete optimization is often a desirable trait. It is especially useful in cases where new data is continuously being added and the model needs to be trained multiple times (e.g. active learning).
\item \textbf{Statistical Efficiency} - Generalizing capacity with increasing the size of the dataset \cite{bojanowski2016enriching}. For e.g. we grab 20\% 40\% 60\% and 80\% of a particular dataset and then train models on them independently. We would expect models to perform better with increase in size of the dataset. The model that performs consistently better (Test losses are almost always the best regardless of the size of the dataset), is loosely considered more efficient.
\end{enumerate}

\subsection{Task}
Character level language modeling is a well-established task for evaluating sequence models~\cite{kalchbrenner2015grid,zilly2016recurrent}. Character level modeling entails predicting the next character token, given you have $k$ previous character tokens.

This is equivalent to estimating the conditional distribution
\begin{eqnarray*}
\mathcal{P}\left( x^{(n)}|\lbrace x^{(n-k)}, \ldots, x^{(n-1)}\rbrace\right)\\
\forall \lbrace x^{(n-k)}, \ldots ,x^{(n-1)}, x^{(n)}\rbrace\in \mathcal{V}^k
\end{eqnarray*}
where $\mathcal{V}$ is the set of all character tokens. All the sequential models in the upcoming experiments are optimized to estimate this conditional distribution.

\subsection{Datasets}
We use Penn Treebank Dataset (henceforth \textbf{\texttt{PTB}}) \cite{marcus2013treebank} with pre-processing in \cite{mikolov2010recurrent} and the War and Peace Dataset (henceforth \textbf{\texttt{WP}}) as the standard benchmarks for character-level language modeling. \texttt{PTB} contains a set of collected 2499 stories designed to allow the extraction of simple predicate and argument structure.
\textbf{\texttt{WP}} is basically the book "War and Peace" by Leo Tolstoy. Being a novel, it brings new challenges to the domain of language modeling because of a huge set of punctuation marks in the data. For example, quotation marks (`` ") come in pairs, forcing the model to learn long range dependencies.
Both these datasets are relatively small and have around 5 million characters. We selected these datasets to represent scenarios with relatively low resources.

Among bigger datasets, we use \textbf{\texttt{enwik8}} and \textbf{\texttt{text8}} from the Hutter Prize dataset \cite{hutter2012}. Both these datasets contain 100 million characters from pages on Wikipedia. While \textbf{\texttt{enwik8}} has XML markups, special characters, latin/non-latin characters adding up to 205 unicode symbols, \textbf{\texttt{text8}} is a much cleaner dataset with only 27 unicode symbols. The shear size of both these datasets is enough to make the task of language modeling challenging. Following common practice, we chose first 90\% for training, next 5\% for validation and last 5\% for testing for all datasets.

\subsection{Training Details}
To make the comparison\footnote{Source code available at: \url{https://github.com/chahuja/lru}} fair, we fixed number of parameters to 10M and 24M based on insights in \cite{collins2016capacity}. For example, all baseline and LRU models will have number of parameters as close as possible to 10 million in 10M experiments. The same is done for the 24M experiments. All models are either 2 or 4 layers deep, except RHNs which were trained with the transition depth of 5 following the protocol in \cite{zilly2016recurrent}.

Batch size was fixed to 250 and all the models are trained by backpropagating the error up till 50 time steps. We use the optimizer Adam\cite{kingma2014adam} with an exponentially (factor of 0.9) decaying learning rate of 0.001, $\beta_1=0.1$ and $\beta_2=0.001$. All weights were initialized using Glorot initialization \cite{glorot2010understanding}. 

\subsection{Evaluation}
We optimize our models with Categorical Cross Entropy (CCE) as the loss function and report the same as part of our evaluation. Lower is better. For the purpose of analysis we store loss values on the held out test set and validation set at every epoch. But the checkpoint with the smallest validation loss is considered our best model, so we report the Test Loss obtained on the best model.


\begin{table*}[ht]
\centering
\caption{This table compares modeling capacity of LSTM, GRU, Grid-LSTM, RHN, PS-LRU, RG-LRU and LRU on the task of character level language modeling. For each dataset, we report losses on the test-split as Categorical Crossentropy Error (CCE) and Time in number of epochs. Lower is better.}
\label{tab:table_10_24_comp}
\begin{tabular}{lcrrrrrrrr}
  \toprule
  &  & \multicolumn{2}{c}{\texttt{PTB}} & \multicolumn{2}{c}{\texttt{WP}} & \multicolumn{2}{c}{\texttt{enwik8}} & \multicolumn{2}{c}{\texttt{text8}}\\
  \cmidrule(lr){3-4}
  \cmidrule(lr){5-6}
  \cmidrule(lr){7-8}
  \cmidrule(lr){9-10}
  & \textbf{Num. of} &\multicolumn{1}{c}{\textbf{Time}}&\multicolumn{1}{c}{\textbf{Loss}}& \multicolumn{1}{c}{\textbf{Time}}&  \multicolumn{1}{c}{\textbf{Loss}} & \multicolumn{1}{c}{\textbf{Time}}&  \multicolumn{1}{c}{\textbf{Loss}}& \multicolumn{1}{c}{\textbf{Time}}&\multicolumn{1}{c}{\textbf{Loss}} \\
\textbf{Models}& \textbf{Params.} &\textbf{(in epochs)}&\textbf{(CCE)}& \textbf{(in epochs)}&  \textbf{(CCE)} & \textbf{(in epochs)}&  \textbf{(CCE)}& \textbf{(in epochs)}&\textbf{(CCE)} \\

  \midrule
  \textbf{LSTM}      &  10M   &       53 &     0.972 &       15 &    1.158 &         52 &      1.075 &          12 &       1.013 \\
  \textbf{GRU}       &  10M   &        9 &     0.972 &       11 &    1.163 &         37 &      1.101 &          17 &       1.047 \\
  \textbf{Grid-LSTM} &  10M   &       10 &     0.970 &       43 &    1.148 &         59 &      1.077 &          15 &       \textbf{1.009} \\
  \textbf{RHN}       &  10M   &       19 &     0.986 &       19 &    1.225 &          6 &      1.206 &          26 &       1.121 \\
  \midrule
  \textbf{PS-LRU}    &  10M   &       8 &     0.971 &        8 &    1.148 &         19 &      1.090 &          17 &       1.019 \\
  \textbf{RG-LRU}    &  10M   &       6 &     0.971 &       11 &    1.143 &         17 &      1.074 &          17 &       1.014 \\
  \textbf{LRU}       &  10M   &       8 &     \textbf{0.965} &        8 &    \textbf{1.141} &         14 &      \textbf{1.057} &          17 &       1.010 \\
  \midrule
  \midrule
  \textbf{LSTM}      &  24M   &        8 &     0.968 &        7 &    1.159 &         11 &      1.035 &          12 &       0.994 \\
  \textbf{GRU}       &  24M   &        9 &     0.980 &        8 &    1.164 &         57 &      1.121 &          59 &       1.089 \\
  \textbf{Grid-LSTM} &  24M   &        7 &     0.971 &        8 &    1.171 &         13 &      \textbf{1.034} &          13 &       \textbf{0.986} \\
  \textbf{RHN}       &  24M  &        18 &     0.979 &       21 &    1.199 &         34 &      1.092 &          18 &       1.108 \\
  \midrule
  \textbf{PS-LRU}    &  24M   &       5 &     0.969 &        5 &    1.152 &         25 &      1.070 &          32 &       1.058 \\
  \textbf{RG-LRU}    &  24M   &       4 &     0.969 &        5 &    1.155 &         23 &      1.065 &          19 &       1.018 \\
  \textbf{LRU}       &  24M   &       5 &     \textbf{0.967} &        6 &    \textbf{1.150} &         17 &      1.041 &          16 &       1.010 \\
\bottomrule
\end{tabular}
\end{table*}

\begin{figure}[ht]
\subfigure[\texttt{PTB}]{\includegraphics[width=0.23\textwidth]{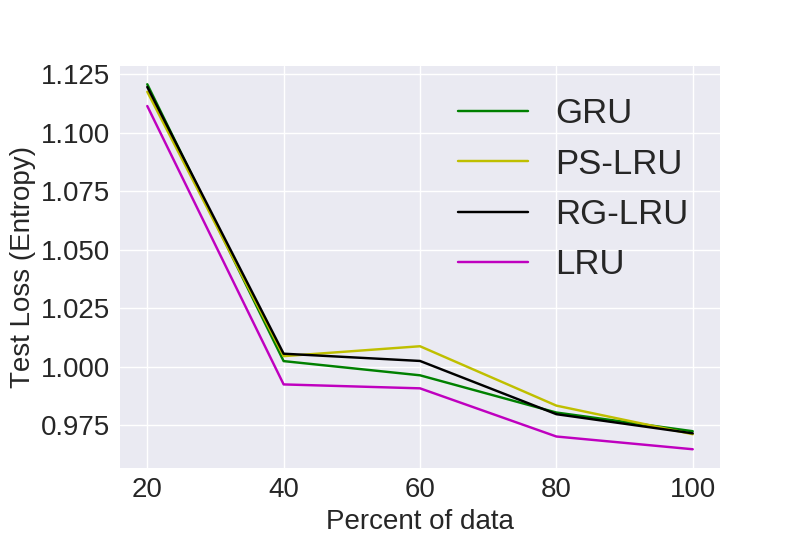}}
\subfigure[\texttt{text8}]{\includegraphics[width=0.23\textwidth]{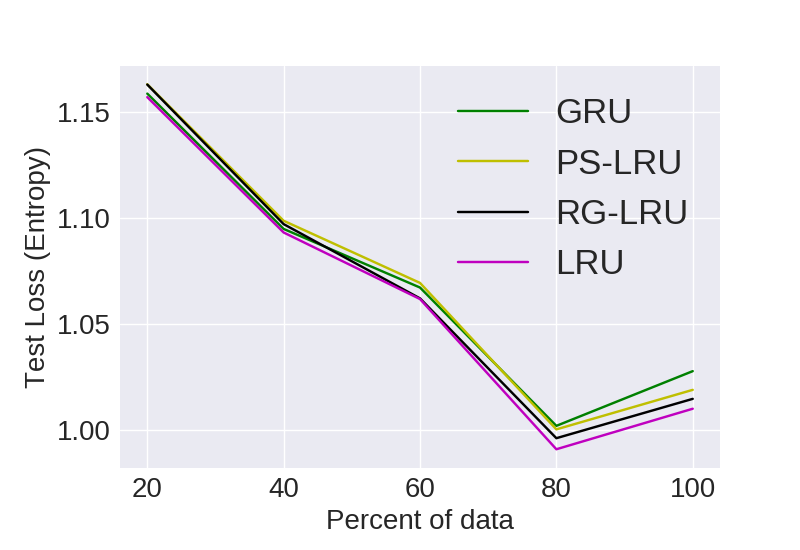}}
\subfigure{\includegraphics[width=0.23\textwidth]{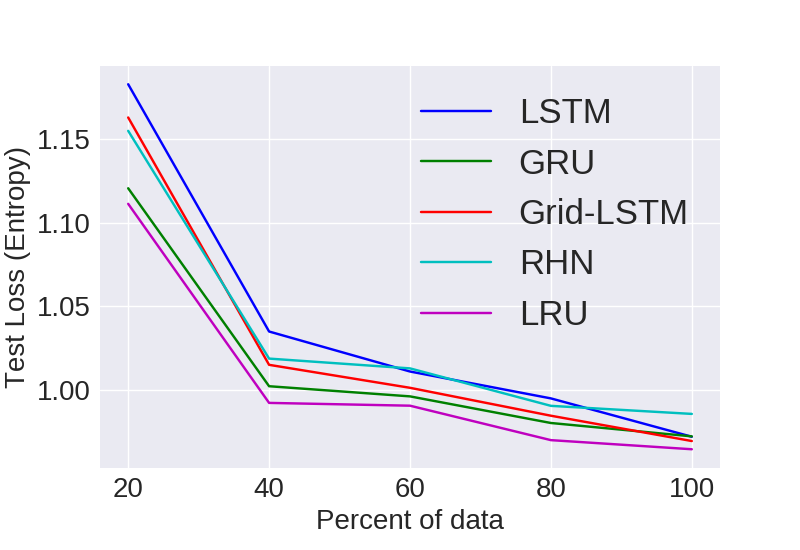}}
\subfigure{\includegraphics[width=0.23\textwidth]{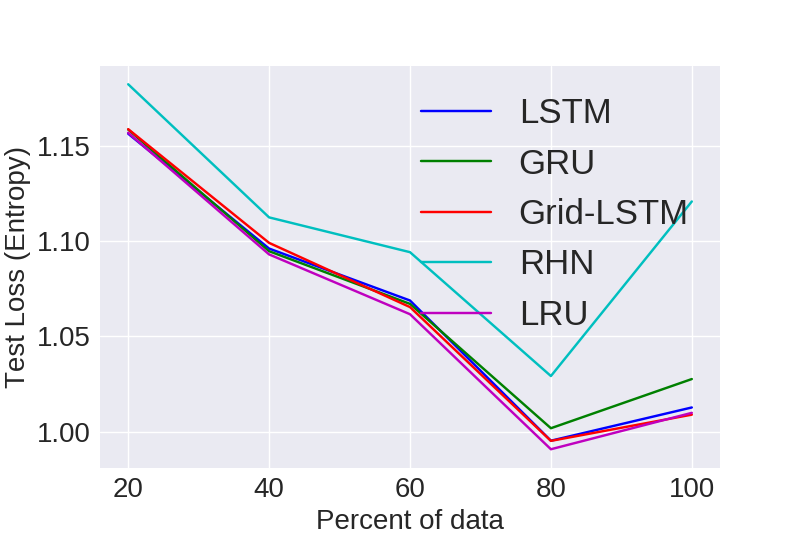}}

\caption{Visual representation of Statistical Efficiency of LSTM, GRU, RHN, Grid-LSTM, PS-LRU, RG-LRU and LRU on \texttt{PTB} and \texttt{text8}. Even though decreasing the amount of data negatively impacts the accuracy of the language model, LRU performs better when compared to other models. A LRU network trained on 40\% of the data is better than other models at 60\% of the data which indicates superiority in low-resource scenarios. Y-axis is test loss for the best set of parameters on validation sets, while X-axis is percent of data used for training. All models have 10M parameters.}
\label{fig:split_curves}
\end{figure}

\footnotetext{Models with 24M parameters were omitted due to space constraints}

\section{Results and Discussion}
\subsection{GRU and LRU Family}
As we mentioned earlier, we tested all recurrent units in the same training environment with parameter budgets of 10 and 24 million parameters. First we compare GRU, PS-LRU, RG-LRU and LRU on all evaluation metrics.

\noindent \textbf{Accuracy:} In Table \ref{tab:table_10_24_comp}, we observe that GRU < PS-LRU < RG-LRU < LRU\footnote{except for \texttt{WP} where PS-LRU is slightly better than RG-LRU} when CCE loss is the validations metric. LRU is consistently the best performing model, and GRU is the worst. It is interesting to note that number of gates is also in the order GRU < PS-LRU < RG-LRU < LRU. This seems to indicate a correlation between the number of gates and performance.

Figure \ref{fig:test_curves_comp} has Test Losses plot against number of epochs. At the end of the first epoch, it seems that the better performing model is already doing better on the held-out set. From then on the nature of the test curve is similar across GRU and LRU family, which prevents the worse model to catch up.

\begin{figure}[!h]
\centering
\includegraphics[width=\linewidth]{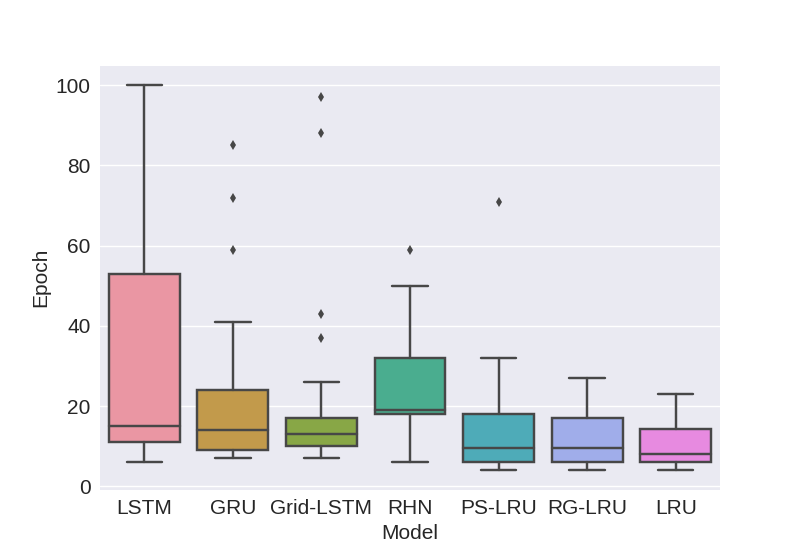}
\caption{Box plot of \textit{number of epochs} required for convergence, across all models and conducted experiments. At an average the family of LRU models converge faster with a low standard deviation in convergence epochs across experiments. LSTM, GRU, Grid-LSTM and RHN networks have a much higher mean and standard deviation in convergence epochs along with many outliers than can take as much as 5 times more than a LRU network to train. Hence, LRU models demonstrate a faster and more stable convergence rate.}
\label{fig:boxplot}
\end{figure}

\noindent \textbf{Computational Convergence Rate:} We also look at the time taken by different networks to converge to the best model (checkpoint at which validation loss is the lowest) expressed in \textit{number of epochs}. The choice of library and recurrent unit implementations\footnote{We use PyTorch implementations of LSTM and GRU which are highly optimized with a C backend in contrast to user-implemented recurrent units} make a huge difference to the speed (per second) of applying gradient descent updates, hence we choose \textit{number of epochs} as the unit for comparison. For a compact overview we consider all\footnote{All experiments include all training instances run on all the models on all datasets discussed in this paper} the experiments we conducted and make a box plot (Figure \ref{fig:boxplot}). It is interesting to note that all variants of LRU, at an average, take around 12 epochs to converge.

\noindent \textbf{Statistical Efficiency:} We grab 20\%, 40\%, 60\% and 80\% of \texttt{PTB} and \texttt{text8} to construct smaller mini-datasets. After running experiments with GRU, and LRU family on the new datasets we calculate the losses on the held-out test set and choose the best models for each mini-dataset. All the models have a parameter budget of 10M. After LRU performed the best amongst its family and GRU, it was not a surprise when it consistently had the best loss for all mini-datasets (Figure \ref{fig:split_curves}). The graphs are an indication of the best generalizing capability and empirical statistical efficiency of LRU (amongst its family and GRU).

\subsection{LRU Family and Baseline Models}
\noindent \textbf{Accuracy:} Next, we compare LSTM, GRU, RHN and Grid-LSTM to LRU in Table \ref{tab:table_10_24_comp}. We see that LRUs perform consistently better than all other recurrent units especially on smaller datasets \texttt{WP} and \texttt{PTB}, which suggests that LRU has better modeling power in scenarios with small training datasets. Figure \ref{fig:test_curves} has Test Losses plot against number of epochs. An interesting behaviour is seen RHNs curves where the test loss rises at first and then starts decreasing, while other models converge almost monotonically to their minimum value. LRU almost always is the best model at the end of one epoch, and it continues to maintain the lead with time.

\noindent \textbf{Computational Convergence Rate:} It is evident from Figure \ref{fig:boxplot}, that Grid-LSTMs, RHNs and GRUs, on an average, take around 22 epochs to converge which is almost double of that of the LRU family. LSTMs, at an average, require even more epoch of around 30 through the complete dataset to converge. Also, LSTMs have a high standard deviation of around 28 epochs. So, LSTMs could potentially take a large number of epochs to converge, while LRUs which have a low standard deviation of around 6 epochs have a more stable convergence rate over different datasets.

\noindent \textbf{Statistical Efficiency:}
As expected, test losses for each model decrease monotonically with the increase in amount of data with an exception of the transition from 80\% to 100\% in \texttt{text8}. One possible reason could be that last 20\% of \texttt{text8} is much harder to model than the rest.
LRUs perform just as good as Grid-LSTMs and LSTMs with increase in the amount of data available for training, if not better. 
For \texttt{PTB}, the difference between the losses of different models increase with the decrease in data (Figure \ref{fig:split_curves}), with the best being LRU and the worst being LSTM. All the models perform equally well on \texttt{text8} with no significant difference across models on its mini-datasets. With the evidence in hand, LRUs seem to have the better empirical statistical efficiency, especially in cases with lesser data. 

\subsection{Vanishing Gradients}
\begin{figure}
\subfigure[GRU]{\includegraphics[width=0.23\textwidth]{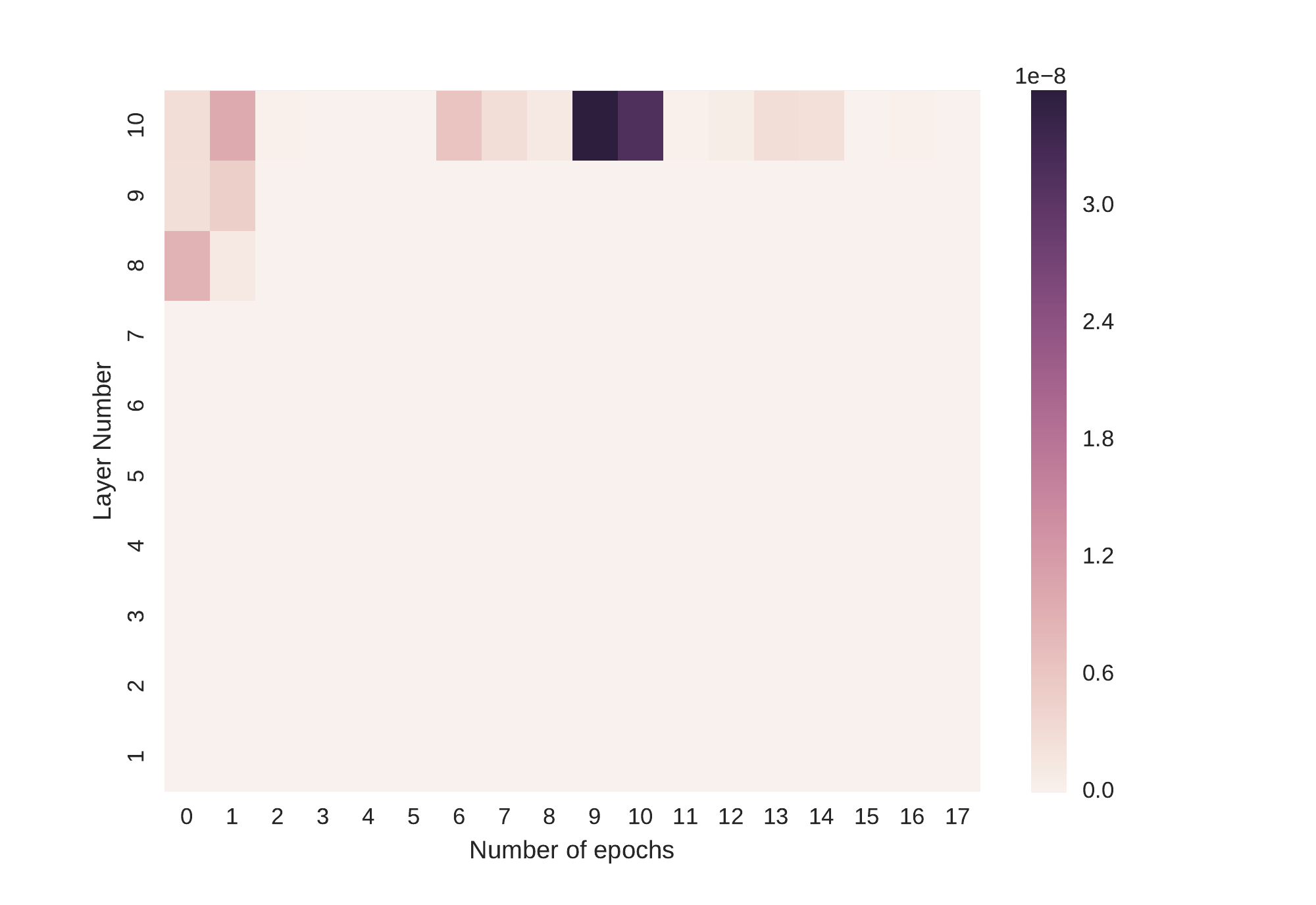}}
\subfigure[LRU]{\includegraphics[width=0.23\textwidth]{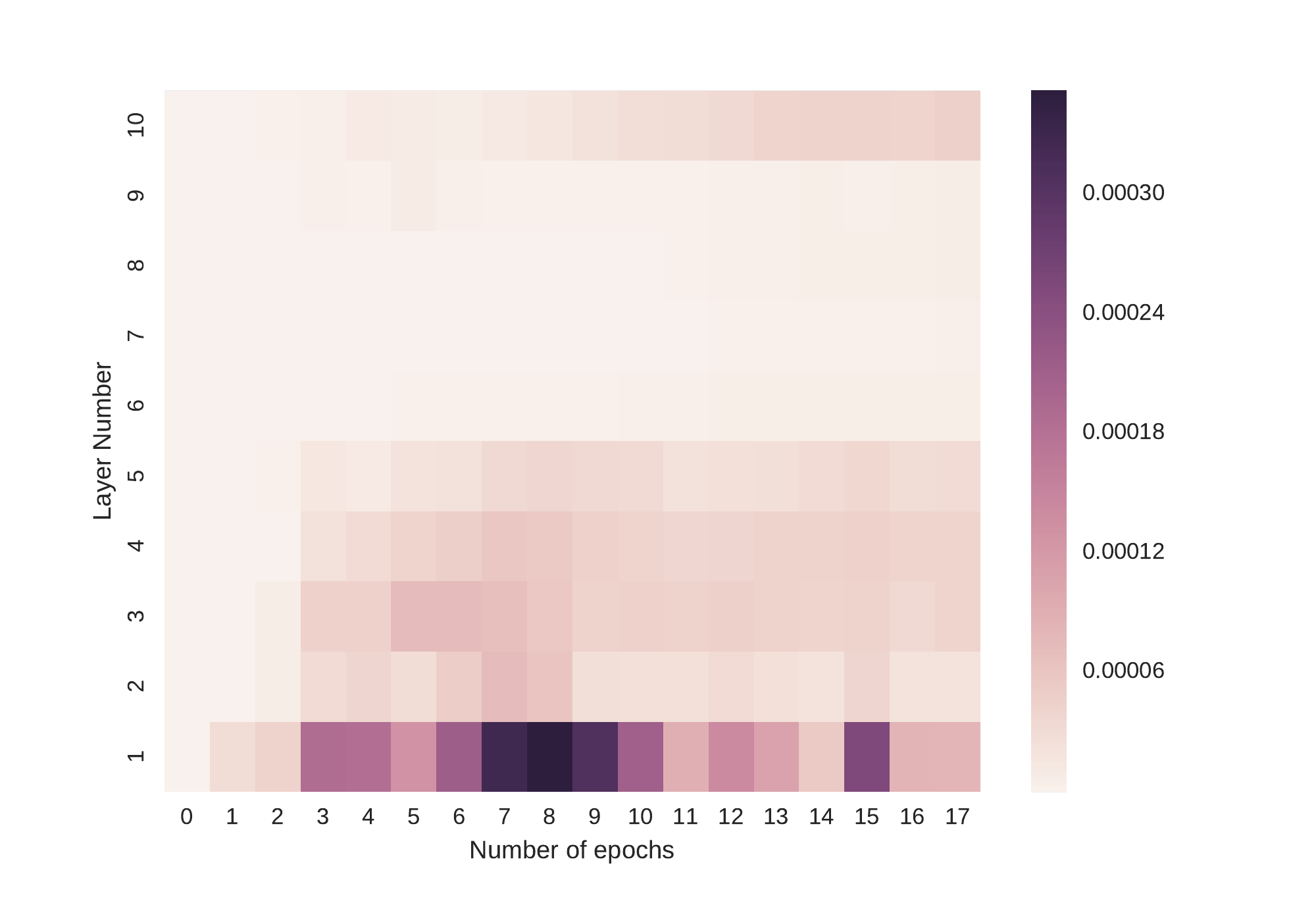}}
\caption{Gradient Norms across layers and number of epochs as a heatmap. It represents the distribution of gradients across layers and number of epochs for each model. Darker values are bigger. Note that the color bar is not the same for both the figures as the mean values of the gradients are far apart from each other. GRUs have extremely low gradients while LRU has high gradients across most layers.}
\label{fig:grads}
\end{figure}

To demonstrate the effectiveness of two distinct streams of information for depth and time, we train a 10 layer deep language model on \texttt{PTB} with LRU and GRU as the recurrent units. As is evident in Figure \ref{fig:grads}, the average gradient norms are concentrated close the last layers in a GRU model, while they are spread across more evenly in a LRU model. This indicates that gradients are backpropagated deep enough for gradient based optimization to work.

\section{Conclusion}
We introduced a new family of models, called Lattice Recurrent Units (LRU), to address the challenge of learning deep multi-layer recurrent models with limited resources.
  Our experiments are performed on four publicly-available datasets, comparing LRU with Grid-LSTM, Recurrent Highway networks, LSTMs and GRUs. Results indicated that LRU has the best accuracy, convergence rate, and statistical efficiency amongst all the baseline models when training language models, especially, when dealing with small datasets.
  We also analyzed the effects of decoupling three different components of our LRU model: \textit{Reset Gate}, \textit{Update Gate} and \textit{Projected State}.
  Amongst GRU, PS-LRU, RG-LRU and LRU, LRU was the best in all the 3 metrics of evaluation mentioned above. In fact, a trend was observed: Decoupling of gates leads to better performance.
Hence, LRU models achieved the goal of learning multi-layer networks with limited resources by creating distinct (but coupled) flow of information along time and depth.

\section{Acknowledgements}
This project was partially supported by Oculus Research Grant. The authors would like to thank Volkan Cirik, Rama Kumar Pasumarthi and Bhuwan Dhingra for their valuable inputs.

\bibliographystyle{aaai}
\bibliography{ref}

\end{document}